\documentclass{article} 
\usepackage{iclr2026_conference,times}

\usepackage{amsmath,amsfonts,bm}

\def\eqref#1{equation~\ref{#1}}

\def\1{\bm{1}}

\DeclareMathAlphabet{\mathsfit}{\encodingdefault}{\sfdefault}{m}{sl}
\SetMathAlphabet{\mathsfit}{bold}{\encodingdefault}{\sfdefault}{bx}{n}

\usepackage{hyperref}
\usepackage{url}
\usepackage{multirow}
\usepackage{graphicx}
\usepackage{booktabs}
\usepackage{multirow}
\usepackage[ruled,vlined]{algorithm2e}
\usepackage{subfigure}
\usepackage{amssymb}

\usepackage[english]{babel}

\title{Deep Self-Evolving Reasoning}

\author{
Zihan Liu\thanks{These authors contributed equally: Zihan Liu, Shun Zheng, Xumeng Wen. Zihan did this work during the internship at Microsoft Research Asia.} $~^1$,
Shun Zheng\footnotemark[1] ~\thanks{Correspondence to shun.zheng@microsoft.com.} $~^2$,
Xumeng Wen\footnotemark[1] $~^2$,
Yang Wang$^2$,
\textbf{
Jiang Bian$^2$,
Mao Yang$^2$} \\
$^1$Peking University
$^2$Microsoft Research Asia
\\
}

\iclrfinalcopy 

\usepackage[most]{tcolorbox} 
\usepackage{xcolor} 
\usepackage{caption}
\usepackage{subcaption}

\newtcolorbox{questionbox}[2][]{
    colback=orange!10, 
    colframe=orange!80!black, 
    colbacktitle=orange!50, 
    title={#2}, 
    sharp corners, 
    boxrule=1pt, 
    titlerule=0pt, 
    fonttitle=\bfseries\color{black}, 
    #1 
}
\newtcolorbox{answerbox}[2][]{
    colback=magenta!10, 
    colframe=magenta!80!black, 
    colbacktitle=magenta!50, 
    title={#2}, 
    sharp corners, 
    boxrule=1pt, 
    titlerule=0pt, 
    fonttitle=\bfseries\color{black}, 
    #1 
}
\newtcolorbox{verifierbox}[2][]{
    colback=green!5!blue!5, 
    colframe=green!40!blue!30, 
    colbacktitle=green!20!blue!10, 
    title={#2}, 
    sharp corners, 
    boxrule=1pt, 
    titlerule=0pt, 
    fonttitle=\bfseries\color{black}, 
    #1 
}

\definecolor{boxblue}{RGB}{220,240,255}
\definecolor{titleblue}{RGB}{50,100,180}
\definecolor{dashblue}{RGB}{100,150,220}

\newtcolorbox{casebox}{
    enhanced,
    colback=boxblue,
    colframe=titleblue,
    arc=4pt,
    boxrule=1.5pt,
    left=10pt,
    right=10pt,
    top=10pt,
    bottom=10pt,
    boxsep=10pt,
}

\newcommand{\sectiontitle}[1]{%
    \vspace{6pt}%
    \noindent\textbf{\color{titleblue}\large #1}%
    \vspace{4pt}%
}

\newcommand{\sectiondivider}{\noindent\rule{\linewidth}{0.4pt}}

\begin{document}

\maketitle

\begin{abstract}
Long chain-of-thought reasoning has become a cornerstone of advanced reasoning in large language models. While recent verification–refinement frameworks have enabled proprietary models to solve Olympiad-level problems, their effectiveness hinges on strong, reliable verification and correction capabilities, which remain fragile in open-weight, smaller-scale models. This work demonstrates that even with weak capabilities for hard tasks, the reasoning limits of such models can be substantially extended through a probabilistic paradigm we call Deep Self-Evolving Reasoning (DSER). We conceptualize iterative reasoning as a Markov chain, where each step represents a stochastic transition in the solution space. The key insight is that convergence to a correct solution is guaranteed as long as the probability of improvement marginally exceeds that of degradation. By running multiple long-horizon, self-evolving processes in parallel, DSER amplifies these small positive tendencies, enabling the model to asymptotically approach correct answers. Empirically, we apply DSER to the \texttt{DeepSeek-R1-0528-Qwen3-8B} model. On the challenging AIME 2024-2025 benchmark, DSER solves 5 out of 9 previously unsolvable problems and boosts overall performance, enabling this compact model to surpass the single-turn accuracy of its 600B-parameter teacher through majority voting. Beyond its immediate utility for test-time scaling, the DSER framework serves to diagnose the fundamental limitations of current open-weight reasoners. By clearly delineating their shortcomings in verification, refinement, and stability, our findings establish a clear research agenda for developing next-generation models with powerful, intrinsic self-evolving capabilities.
\end{abstract}

\begin{figure*}[h]
\centering
\includegraphics[width=\textwidth]{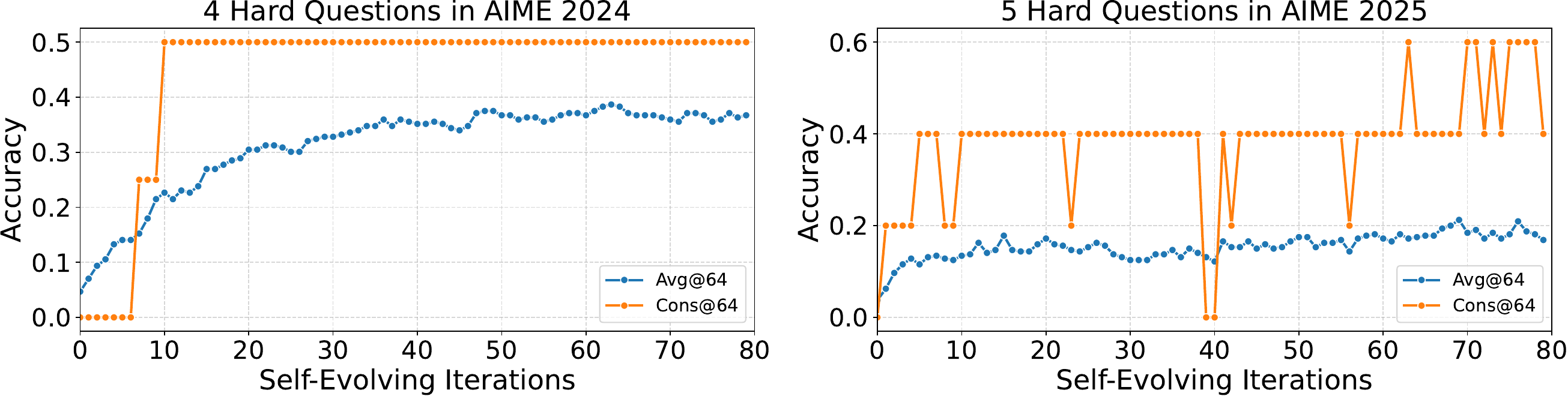}
\caption{
Deep self-evolving reasoning enables \texttt{DeepSeek-R1-0528-Qwen3-8B} to solve 5 of 9 AIME 2024-2025 problems previously deemed ``unsolvable'' by standard majority voting over parallel trials (Avg@64: average accuracy over $64$ runs, Cons@64: consistency accuracy over $64$ runs).
A notable example is the success for a difficult problem in AIME 2025 after $80$ self-evolving iterations, a process consuming approximately 10 million reasoning tokens. The final correct answer can be determined by a majority vote across the last ten self-evolving iterations.
}
\label{fig:acc_on_aime_hard}
\end{figure*}

\section{Introduction}
\label{sec:intro}

Chain-of-Thought (CoT) reasoning~\citep{wei2022CoT}, a cornerstone technique in large language models (LLMs), has driven rapid progress in advancing reasoning capability.
It was first demonstrated in OpenAI’s o1~\cite{openai2024o1} series models that increasing the length of CoT directly leads to test-time scaling, enabling LLMs to tackle more complex and challenging tasks.
Following this, DeepSeek-R1~\citep{guo2025deepseek-r1-nature} became the first open-source effort to realize long-form CoT reasoning through reinforcement learning. 
At the heart of this approach lies the Group Relative Policy Optimization (GRPO) algorithm~\citep{shao2024deepseekmath}, which effectively incentivizes high-quality reasoning traces in pre-trained LLMs~\citep{wen2025RLVR-correct}.
Since the public release of DeepSeek-R1, the community has witnessed a wave of reproductions~\citep{yu2025DAPO-Zero,he2025Skywork-OR1,liu2025DrGRPO-Oat-Zero,hu2025OpenReasonerZero} and the surge of large-scale, high-performance reasoning models~\citep{yang2025Qwen3,kimi2025K2,zeng2025GLM-4.5} in the open-source ecosystem.

Building on long CoT reasoning, frontier industry labs claimed advanced reasoning systems whose performance rivals that of IMO 2025 gold medalists~\citep{OpenAI2025IMO-Gold,Gemini2025IMO-Gold,chen2025Seed-Prover}.
An independent study~\citep{huang2025IMO-Gold-Agent} further reported that state-of-the-art proprietary models, such as Gemini 2.5 Pro~\citep{Gemini2025IMO-Gold}, GPT-5~\citep{openai2025GPT-5}, and Grok-4~\citep{xai2025Grok-4}, can solve 5 out of 6 IMO problems using a model-agnostic, verification–refinement framework. 
While similar self-refining concepts had already emerged in prior studies~\citep{kim2023RecurCritImpr,madaan2023self-refine,kamoi2024survey-self-correct,kumar2024self-correct-RL,bensal2025reflect-retry-reward}, this framework offered concrete and practical insights, clearly demonstrating the immense potential of iterative reasoning calls to solve problems at the IMO level.

However, the framework introduced in~\citep{huang2025IMO-Gold-Agent} relies heavily on advanced verification, refinement, and instruction-following abilities, which remain largely exclusive to leading proprietary models when handling extremely hard reasoning tasks.
It is still unclear to what extent open-weight reasoning models, especially small and medium-sized ones with broader accessibility, can benefit from self-evolving paradigms and extend their reasoning limits.
In practice, such models often exhibit weak self-verification, occasional self-improvement, and unstable instruction-following behaviors, leading to unexpected terminations under~\citeauthor{huang2025IMO-Gold-Agent}’s framework.

In this work, we show that even when a model exhibits weak verification and refinement capabilities on hard reasoning tasks, a simple self-evolving setup with concise prompts could still substantially extend the reasoning boundary.
Our approach begins from a probabilistic interpretation of the classic verification–refinement iteration: we view each iteration as a transition step of a self-evolving stochastic process.
The model’s verification and refinement abilities determine the transition probability matrix for a given problem, forming a Markov chain whose convergence can be theoretically guaranteed.
As long as the probability of improvement (transitioning from an incorrect to a correct solution) exceeds the probability of degradation (from correct to incorrect), the process converges to a stationary distribution dominated by correct solutions.
By running multiple independent self-evolving processes over sufficiently long iterations, the model can fully unlock its inherent self-evolving potential.
We refer to this general paradigm as Deep Self-Evolving Reasoning (DSER).

The core insight of DSER lies in its probabilistic view of self-improvement.
Rather than expecting each round of verification and refinement to succeed with high accuracy, DSER leverages the convergence property of Markov chains to ensure asymptotic correctness.
It treats multi-turn reasoning as a stochastic optimization trajectory in the discrete token space, where small but statistically positive tendencies toward improvement are sufficient to guarantee long-term convergence toward correct solutions.
In practice, we observe that even when the degradation probability exceeds the improvement probability, parallel DSER procedures could still produce a correct majority-voting answer because correct solutions converge to the same ground-truth while incorrect ones diverge in different results.
Moreover, we note that any verification-refinement iterations can be viewed as a self-evolving stochastic process, including~\citeauthor{huang2025IMO-Gold-Agent}'s framework.
The key distinction is that they allocated more reasoning budgets to verification and add specific conditions to exit the loop.

We evaluate our approach using \texttt{DeepSeek-R1-0528-Qwen3-8B}, configured with up to $64$K response tokens per reasoning call.
Although this model exhibits strong reasoning ability for its scale, it fails to solve $9$ problems (under majority voting) out of $60$ in AIME 2024 and 2025 benchmarks.
For these challenging cases, the average Pass@1 is below 0.05, and both verification and correction success rates remain low. 
As shown in Figure~\ref{fig:acc_on_aime_hard}, applying DSER enables the model to solve $5$ of these $9$ hard problems through majority voting.
Notably, this includes one problem with an initial single-turn Pass@1 of zero (estimated over 128 samples).
These results indicate that DSER successfully extends the single-turn reasoning boundaries of this 8B model.
Moreover, our additional experiments show that when applied to the entire AIME benchmark, DSER improves its Pass@1 accuracy by 6.5\% on AIME 2024 and by 9.0\% on AIME 2025.
Specifically, DSER enables the majority-voting accuracy of this 8B model to surpass the Pass@1 performance of its 600B-parameter teacher model, \texttt{DeepSeek-R1-0528}. This demonstrates that DSER effectively trades test-time computation for enhanced model capacity.

The implications of this work extend beyond its core demonstration of self-evolution under imperfect verification and refinement. For instance, the approach could improve the exploration stage in GRPO training, helping to uncover successful reasoning pathways for extremely difficult problems. Moreover, it could help to reduce the deployment cost while maintaining comparable reasoning performance. Furthermore, our experimental results also reveal significant shortcomings in existing open-weight reasoning models. A key direction for future research is therefore to develop models that are capable of problem-solving, self-verification, providing constructive feedback, increasing correction likelihood, avoiding potential degradation, etc.

\section{Related Work}
\label{sec:rel_work}

Iterative verification and refinement has emerged as a foundational technique for enhancing the reasoning capabilities of LLMs, appearing under various names in the literature. Early work explored this concept through frameworks for recursive self-critique and improvement~\citep{kim2023RecurCritImpr}, as well as using a single model to generate, refine, and provide feedback on its own outputs~\citep{madaan2023self-refine}. This line of research encompasses related ideas such as self-correction~\citep{kumar2024self-correct-RL} and self-verification or self-reflection~\citep{weng2022self-verify-better-reason, bensal2025reflect-retry-reward}. The effectiveness of these methods, however, often depends on the quality of feedback. As noted by \citet{kamoi2024survey-self-correct}, self-correction is most successful when guided by reliable external signals—a principle dramatically demonstrated by systems like Seed-Prover~\citep{chen2025Seed-Prover}, which achieved state-of-the-art performance on IMO 2025 problems by integrating iterative reasoning with formal verification. Concurrently, \citet{huang2025IMO-Gold-Agent} showed that a sophisticated verification-refinement pipeline could enable leading proprietary models to solve problems at an IMO gold medal level.

A considerable body of recent research has focused on endowing LLMs with more robust, intrinsic capabilities for self-verification and self-improvement through specialized training objectives~\citep{kumar2024self-correct-RL, bensal2025reflect-retry-reward, yuan2025Agent-R}. Another related direction involves multi-turn tool use, which can be viewed as a form of iterative refinement guided by external tools and environments~\citep{feng2025retool, dong2025ARPO, shang2025rstar2-agent}. These developments reflect a broader research trend toward self-evolution in LLMs, a paradigm shift that extends beyond single-turn reasoning to more powerful capabilities~\citep{tao2024survey-self-evolution,huang2025R-Zero}.

Our work distinguishes itself by introducing a novel, probabilistic interpretation of the verification-refinement loop. We conceptualize iterative reasoning as a stochastic process governed by a Markov chain. This formulation provides a theoretical basis for improvement even when the model's verification and refinement capabilities are imperfect (typical conditions for hard tasks), as the process can converge to a correct solution given a marginal statistical bias towards improvement. This perspective allows LLMs to progressively solve previously intractable problems and reliably uncover effective reasoning pathways, advancing the frontier of what is achievable with open-weight models.

\section{Methodology}
\label{sec:method}

\begin{figure*}[t]
\centering
\includegraphics[width=\textwidth]{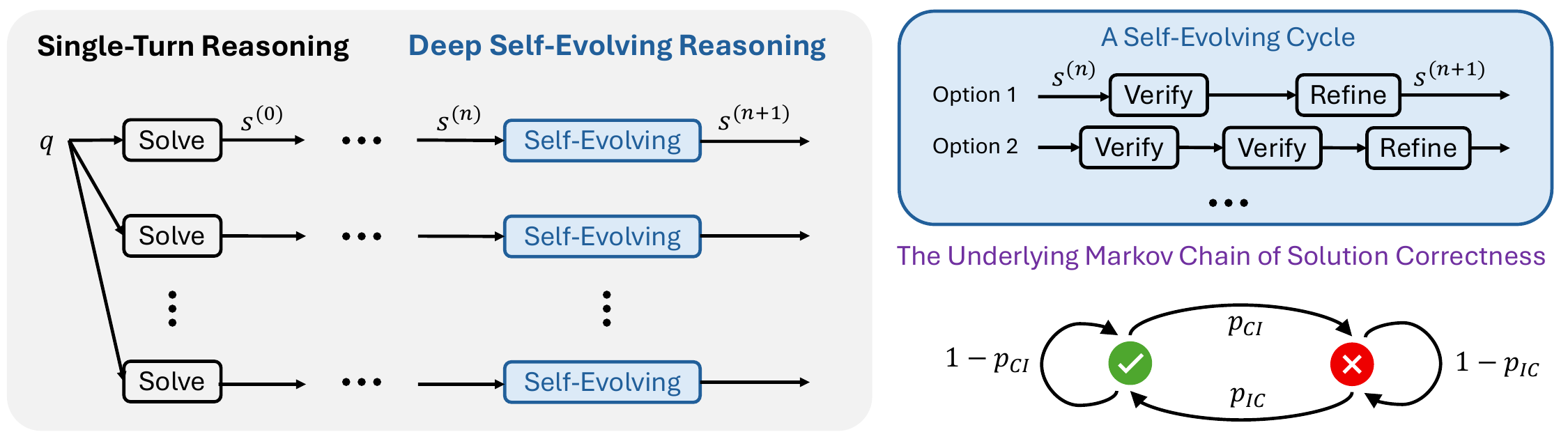}
\caption{
An overview of our DSER approach, where each rectangle of ``Solve'', ``Verify'', and ``Refine'' corresponds to one LLM reasoning call. In the view of Markov chain, a sufficient condition to elicit correct solutions for hard problems is to self-evolve deeply.
}
\label{fig:dser_framework}
\end{figure*}

Our approach models the iterative verification and refinement of solutions as a self-evolving stochastic process. This probabilistic framework allows us to analyze the trajectory of a solution's quality and understand its convergence towards correctness.
Figure~\ref{fig:dser_framework} gives an overview of our approach. 

\begin{figure*}[t]
\centering
\includegraphics[width=\textwidth]{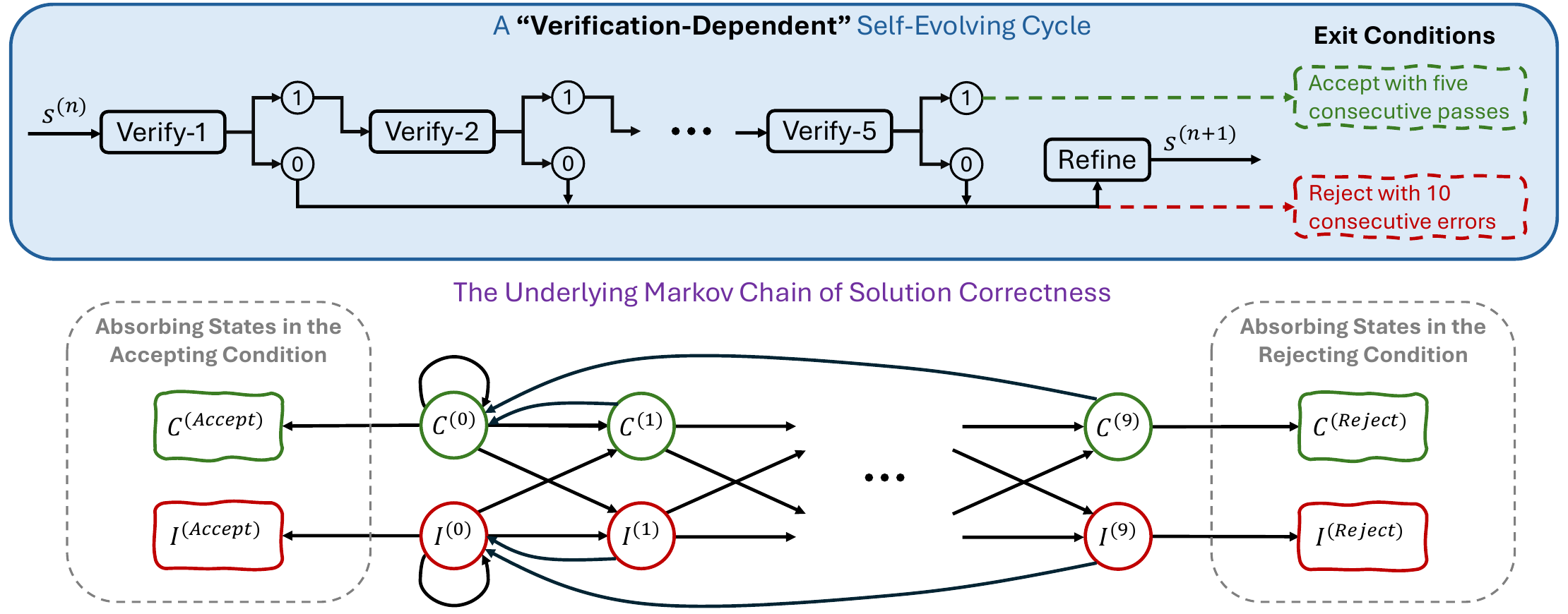}
\caption{
Through the lens of Markov chain, we revisit the iterative verification-refinement cycle proposed by~\citeauthor{huang2025IMO-Gold-Agent}. Here we need to define multiple states of solution correctness indexed by the number of consecutive self-verified rejections and refinements. For instance, $C^{(9)}$ denotes the solution being correct after $9$ consecutive rounds of self-verified rejections and refinements. 
}
\label{fig:verify_dependent_revisit}
\end{figure*}

\subsection{The Self-Evolving Process}
\label{sec:method_self-evolving_proc}

Given an initial question prompt $q$, a reasoning LLM generates a candidate solution $s$. This initial step can be formally represented as:
\begin{align}
s = \mathcal{R}^{LLM}(q),
\label{eq:solve}
\end{align}
where $\mathcal{R}^{LLM}(\cdot)$ denotes a reasoning call that takes a prompt and outputs a summarized solution after long CoT thinking. We define this initial solution as $s^{(0)}$.

The process then enters a series of self-evolving iterations, where the solution at iteration $n$, denoted by $s^{(n)}$, is transformed into $s^{(n+1)}$ through a cycle of self-improvement. In this iteration, various verification-refinement interaction schemes are possible.
Below we present the fundamental two-step cycle as an example. First, a verification step provides feedback on the current solution. Let $p_v$ be the verification prompt designed to elicit this feedback. The resulting verification output $v^{(n)}$ is generated as:
\begin{align}
v^{(n)} = \mathcal{R}^{LLM} ([q; s^{(n)}; p_v]),
\label{eq:verify}
\end{align}
where $[q; s^{(n)}; p_v]$ denotes the concatenation of the original question, the current solution, and the verification prompt as context for the LLM.

Next, a refinement step uses this feedback to generate an improved solution. Let $p_r$ be the refinement prompt. The next-state solution $s^{(n+1)}$ is produced by:
\begin{align}
s^{(n+1)} = \mathcal{R}^{LLM} ([q; s^{(n)}; p_v; v^{(n)}; p_r]).
\label{eq:refine}
\end{align}
This iterative process, transforming $s^{(n)} \to s^{(n+1)}$, continues until a termination condition is met, such as a fixed number of iterations.

\subsection{Markov Chain Formulation}
\label{sec:method_markov}

To analyze the dynamics of this process, we model the evolution of the solution's correctness as a Markov chain. Let us define a discrete state space $\mathcal{S} = \{C, I\}$, where $C$ denotes that the solution $s^{(n)}$ is ``Correct'' and $I$ denotes that it is ``Incorrect''. Let $X_n$ be the random variable representing the state of the solution at iteration $n$. The evolution of the system is then described by the distribution over these states.

According to Equations~\ref{eq:verify},~\ref{eq:refine}, the correctness of the next solution $s^{(n+1)}$ depends on the correctness of the current solution $s^{(n)}$ and not on the history of previous solutions $\{s^{(0)}, \dots, s^{(n-1)}\}$.
Moreover, we assume the improvement capability of the LLM is consistent across self-evolving iterations for a given problem.
Thus a single \textbf{transition probability matrix} $P$ governs this evolution.
\begin{align}
P = 
\begin{pmatrix}
1 - p_{CI} & p_{CI} \\
p_{IC} & 1 - p_{IC}
\end{pmatrix}
=
\begin{pmatrix}
P(X_{n+1}=C | X_n=C) & P(X_{n+1}=I | X_n=C) \\
P(X_{n+1}=C | X_n=I) & P(X_{n+1}=I | X_n=I)
\end{pmatrix}
\label{eq:trans_prob}
\end{align}
where:
\begin{itemize}
    \item $p_{IC}$ is the probability of \textbf{improvement} (moving from Incorrect to Correct).
    \item $p_{CI}$ is the probability of \textbf{degradation} (moving from Correct to Incorrect).
\end{itemize}
The specific values of $p_{IC}$ and $p_{CI}$ depend on the capability of the LLM on solving the problem $q$.
Notably, in this formulation, we do not rely on the accuracy of each verification or refinement reasoning call.
As long as the LLM has some chances to improve towards the correct solution, the transition matrix will guide the evolution towards a stationary distribution.

\subsection{Stationary Distribution and Convergence}
\label{sec:method_distr_converge}

For an ergodic Markov chain (which holds if $p_{IC} > 0$ and $p_{CI} > 0$), the process will converge to a unique stationary distribution $\pi = [\pi_C, \pi_I]$, which satisfies the equation $\pi P = \pi$. This distribution represents the long-term probability of the solution being in either state.

Solving $\pi P = \pi$ subject to the constraint $\pi_C + \pi_I = 1$, we get the stationary probabilities:
\begin{align}
\pi_C = \frac{p_{IC}}{p_{IC} + p_{CI}} \quad \text{and} \quad \pi_I = \frac{p_{CI}}{p_{IC} + p_{CI}}.
\label{eq:station_pi}
\end{align}

\paragraph{Robustness to Imperfect Verification and Refinement for Hard Problems}
Equation~\ref{eq:station_pi} tells us that as long as $p_{IC} > p_{CI}$, meaning the tendency of LLM to improve overweigh that to degrade, running self-evolving iterations sufficiently long will guide the convergence towards a state where the majority of solutions are correct. This gives a theoretical guarantee for the majority voting of parallel DSER processes. And we do not depend on the success of single verification or refinement steps. In practice, we find that even when $p_{IC} < p_{CI}$ for some very hard problems beyond the LLM's existing capabilities, as long as $p_{IC}$ is not too small, the majority voting of multiple DSER processes could still be correct because all correct solutions arrive at the same ground truth while different incorrect solutions diverge in different ways with inconsistent answers.

\paragraph{Convergence Speed}
The speed of convergence to this stationary distribution is determined by the second-largest eigenvalue in magnitude of the transition matrix $P$, given by $|\lambda_2| = |1 - p_{CI} - p_{IC}|$.  In an ideal scenario where $p_{CI} \rightarrow 0$ and $p_{IC} \rightarrow 1$, indicating the LLM consistently corrects errors without degrading correct solutions, the stationary distribution converges to $\pi_C \rightarrow 1$. This scenario, typical for easy problems, yields extremely fast convergence as $|\lambda_2| \rightarrow 0$. For more challenging problems, the improvement probability $p_{IC}$ is often small. However, if the LLM can maintain a correct solution with high probability (i.e., $p_{CI}$ is also small), then $|\lambda_2| = 1 - p_{IC} - p_{CI} < 1$, still guaranteeing exponential convergence at a rate of $|\lambda_2|^n$ over $n$ iterations.

\paragraph{Reinforcement Learning for Self-Evolving Reasoning}
Our approach also delivers unique insights informing future reinforcement learning designs for self-evolving reasoning.
For instance, in addition to purely optimizing self-verification or self-correction capabilities~\citep{bensal2025reflect-retry-reward,kumar2024self-correct-RL}, we could develop new optimization objectives to improve $p_{IC}$ and decrease $p_{CI}$ explicitly.
Moreover, we could integrate the idea of deep self-evolving into the exploration stage to identify more possible solutions for hard tasks.

\subsection{Comparison with Verification-Dependent Self-Evolving}
\label{sec:method_verify-depend}

Our probabilistic perspective also allows us to reinterpret the framework of \citeauthor{huang2025IMO-Gold-Agent} as a self-evolving process. The upper part of Figure~\ref{fig:verify_dependent_revisit} illustrates the core operations in their self-evolving cycle. The key distinction lies in the cycle's dependence on self-verification outcomes (Pass: 1, Fail: 0). The process reaches an accepting condition after five consecutive self-verified passes, deeming the current solution correct. Conversely, it triggers a rejecting condition after ten consecutive verification failures, concluding that the problem is unsolvable by the framework. Given its heavy reliance on verification feedback, we term this a ``verification-dependent'' self-evolving process.

We analyze the underlying Markov chain of this process, depicted in the lower part of Figure~\ref{fig:verify_dependent_revisit}. The verification-dependent design necessitates numerous states to track the count of consecutive rejections. The chain reaches absorbing states when either condition is met: the rejecting condition after ten consecutive failures, or the accepting condition after five consecutive passes. Any single verification pass resets the rejection counter.

Crucially, these verification-induced absorbing states can hinder deep self-evolution for open-weight models on hard problems. The rejecting condition prematurely terminates exploration when the model is perplexed, while the accepting condition risks cementing a false-positive solution. In practice, we find the former limitation more constraining. Furthermore, the rejecting condition renders the Markov chain analytically intractable.

Even without the rejecting condition, the framework remains verification-dependent due to the accepting condition. This simplified Markov chain, with only four states, becomes amenable to theoretical analysis. Our analysis (detailed in Appendix~\ref{supp_sec:theory_verify_depend}) confirms that achieving a favorable stationary distribution requires non-trivial assumptions about self-verification accuracy—assumptions that often fail for hard problems beyond the model's current capability.

In contrast, our DSER framework marginalizes over the verification outcome, relying solely on the relative strength of improvement versus degradation tendencies. This fundamental difference suggests that deep self-evolving, by circumventing the need for precise verification, offers a more viable path for open-weight models to narrow the performance gap with leading proprietary systems.

\section{Experiments}
\label{sec:exp}

We apply our DSER approach to \texttt{DeepSeek-R1-0528-Qwen3-8B} (abbreviated as \texttt{DS-8B}), a powerful 8B-parameter reasoning LLM distilled from a 600B teacher model. We follow its standard inference setup\footnote{\url{https://huggingface.co/deepseek-ai/DeepSeek-R1-0528-Qwen3-8B}}, allowing up to $64K$ response tokens per reasoning call. We use AIME 2024 and 2025, totaling $60$ mathematical competition problems, as our evaluation benchmarks.

Despite its strong baseline performance, \texttt{DS-8B} failed to solve $9$ of these problems. We classify these $9$ problems as "unsolvable" by the base reasoning model, as it could not produce a correct solution even with majority voting over $128$ parallel trials. Additionally, we apply DSER to the entire AIME 2024-2025 problem set to demonstrate its overall performance improvement.

We run $K$ independent DSER trials for each problem and report two metrics:
\begin{itemize}
    \item \textbf{Average Accuracy (Avg@$K$)}: The average accuracy across the $K$ trials. This estimates the Pass@$1$ success probability of a single reasoning process.
    \item \textbf{Consistency Accuracy (Cons@$K$)}: The accuracy of the single solution derived from a majority vote (consistency prediction) over the $K$ trial outputs. This estimates the majority-voting performance over parallel reasoning processes.
\end{itemize}

We employ concise prompts designed to elicit the model's inherent verification and refinement capabilities. In addition to the vanilla problem-solving prompt, our self-evolving stage utilizes the following specialized prompts.

\textbf{Verification Prompt:}
\begin{verbatim}
Verify the given solution step by step to check correctness. 
Provide a short verification report, containing the key points 
of the solution and any errors found. Finally, put your 
judgement strictly in the format: \boxed{1} if correct, 
or \boxed{0} if incorrect.
\end{verbatim}

\paragraph{Refinement Prompt:}
\begin{verbatim}
Given your previous solution and verification report, reconsider
the problem carefully and provide a corrected solution.
Output your final answer strictly in the format: \\boxed{}.
\end{verbatim}

\paragraph{Extended Reasoning Limit for \texttt{DS-8B}}
As Figure~\ref{fig:acc_on_aime_hard} shows, our DSER approach unlocks latent reasoning capabilities in \texttt{DS-8B}, enabling it to solve hard problems that are intractable with its baseline single-turn reasoning paradigm.
Simultaneously, we observe that convergence to the stationary distribution can be slow, as indicated by the steady improvement of Avg@64 even after $80$ iterations.
However, the majority-voting performance (Cons@K) increases rapidly within the first ten iterations for most problems that DSER ultimately solves.
These observations align with our Markov chain perspective, where iterative verification and refinement are modeled as a stochastic process.
Thus, the convergence of solution correctness for a specific problem depends on the model's probabilities of improving versus degrading its solution.
The slow convergence indicates that these problems are exceptionally difficult for \texttt{DS-8B}, implying a small corresponding improvement probability $p_{IC}$.

\begin{figure*}[t]
\centering
\includegraphics[width=\textwidth]{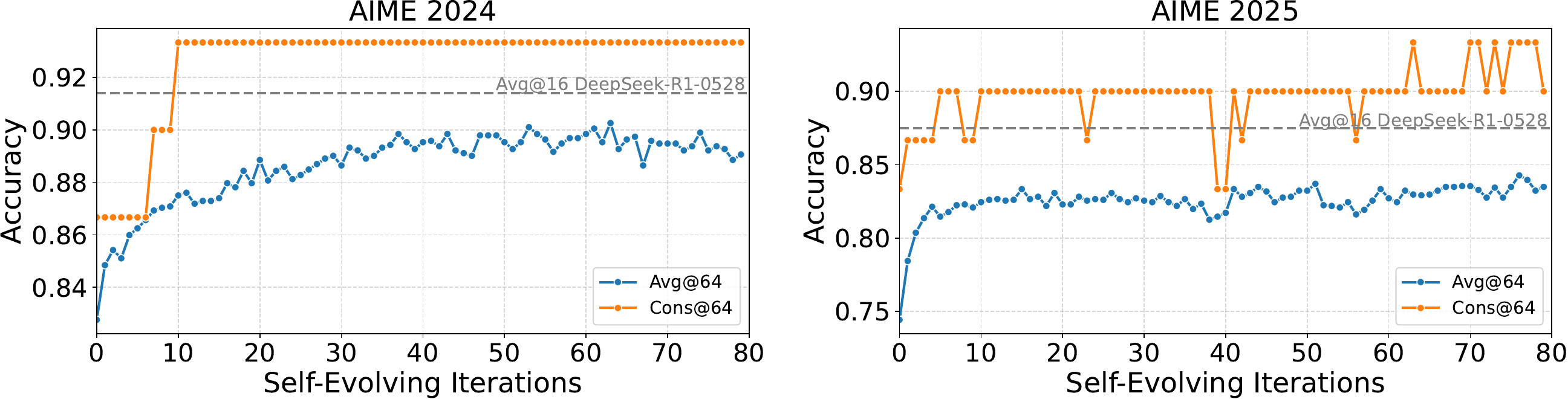}
\caption{
Overall performance of \texttt{DS-8B} with DSER over iterations on the full AIME 2024 and 2025 benchmarks. We specifically flag the Avg@16 metric reported for \texttt{DeepSeek-R1-0528}, which is the 600B distillation teacher model for \texttt{DS-8B}.
}
\label{fig:acc_on_aime_all}
\end{figure*}

\paragraph{Overall Improved Performance}
In addition to solving previously unsolvable problems, Figure~\ref{fig:acc_on_aime_all} shows that DSER stably improves the overall performance of \texttt{DS-8B} across the entire AIME benchmark.
While the breakthrough in majority-vote accuracy (Cons@64) is primarily driven by solving these hard problems, DSER also boosts the overall Pass@1 performance (Avg@64) for all questions: improving from 82.8\% to 89.3\% on AIME 2024 (+6.5\%), and from 74.4\% to 83.4\% on AIME 2025 (+9.0\%).
These results demonstrate that our DSER approach effectively translates the test-time scaling of \texttt{DS-8B} into improved reasoning capacity.
Notably, a small gap remains between the converged Pass@1 performance of \texttt{DS-8B} and its 600B teacher model. This indicates that the stationary distribution of the 8B model's self-evolution is still weaker than the single-turn reasoning capacity of \texttt{DeepSeek-R1-0528}.

\begin{figure*}[ht]
\centering
\includegraphics[width=\textwidth]{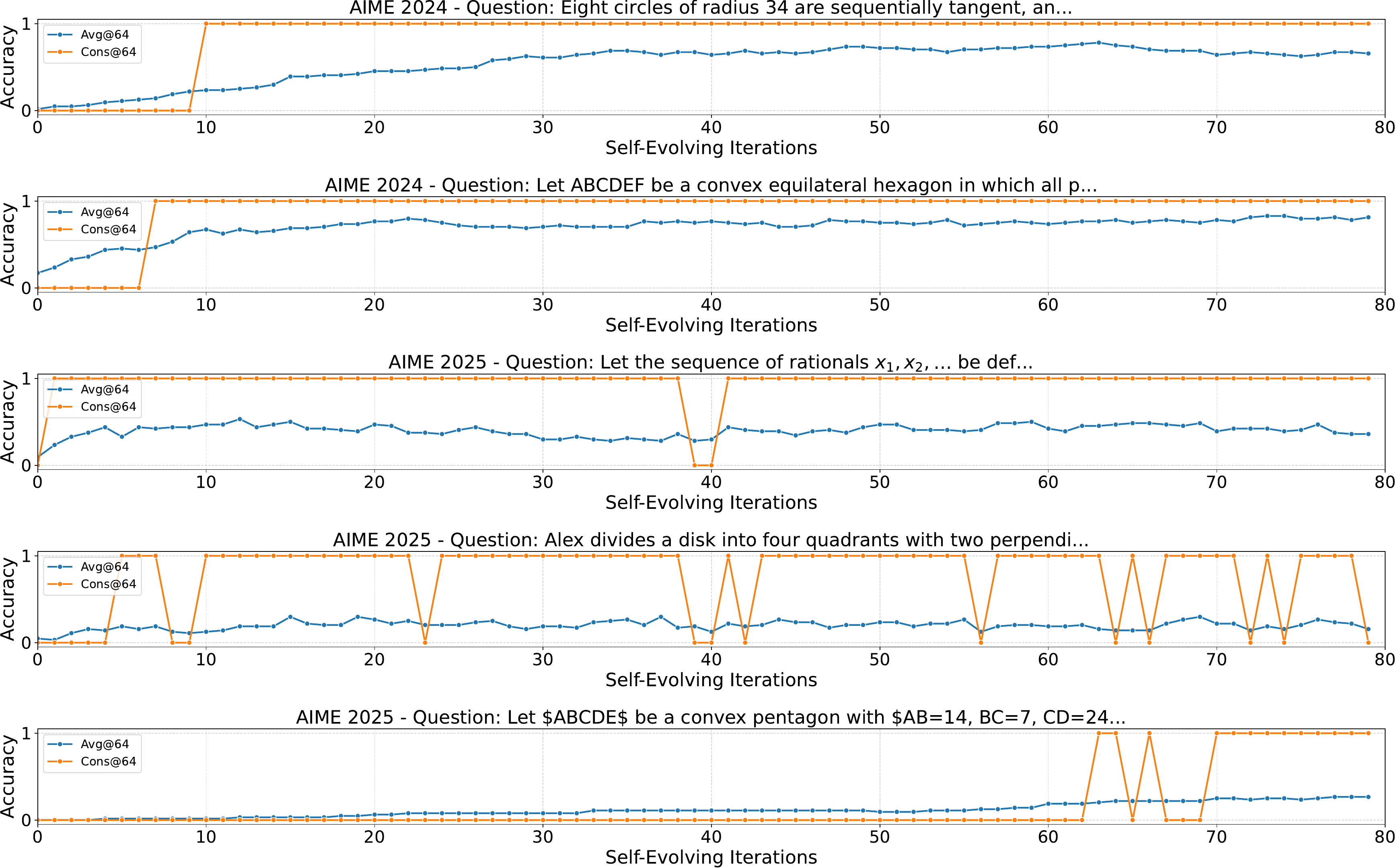}
\caption{
Per-question performance improvements on hard problems over self-evolving iterations, highlighting the diverse convergence speeds and stationary distributions of solution correctness.
}
\label{fig:acc_on_aime_hard_per_q}
\end{figure*}

\begin{figure*}[ht]
\centering
\includegraphics[width=\textwidth]{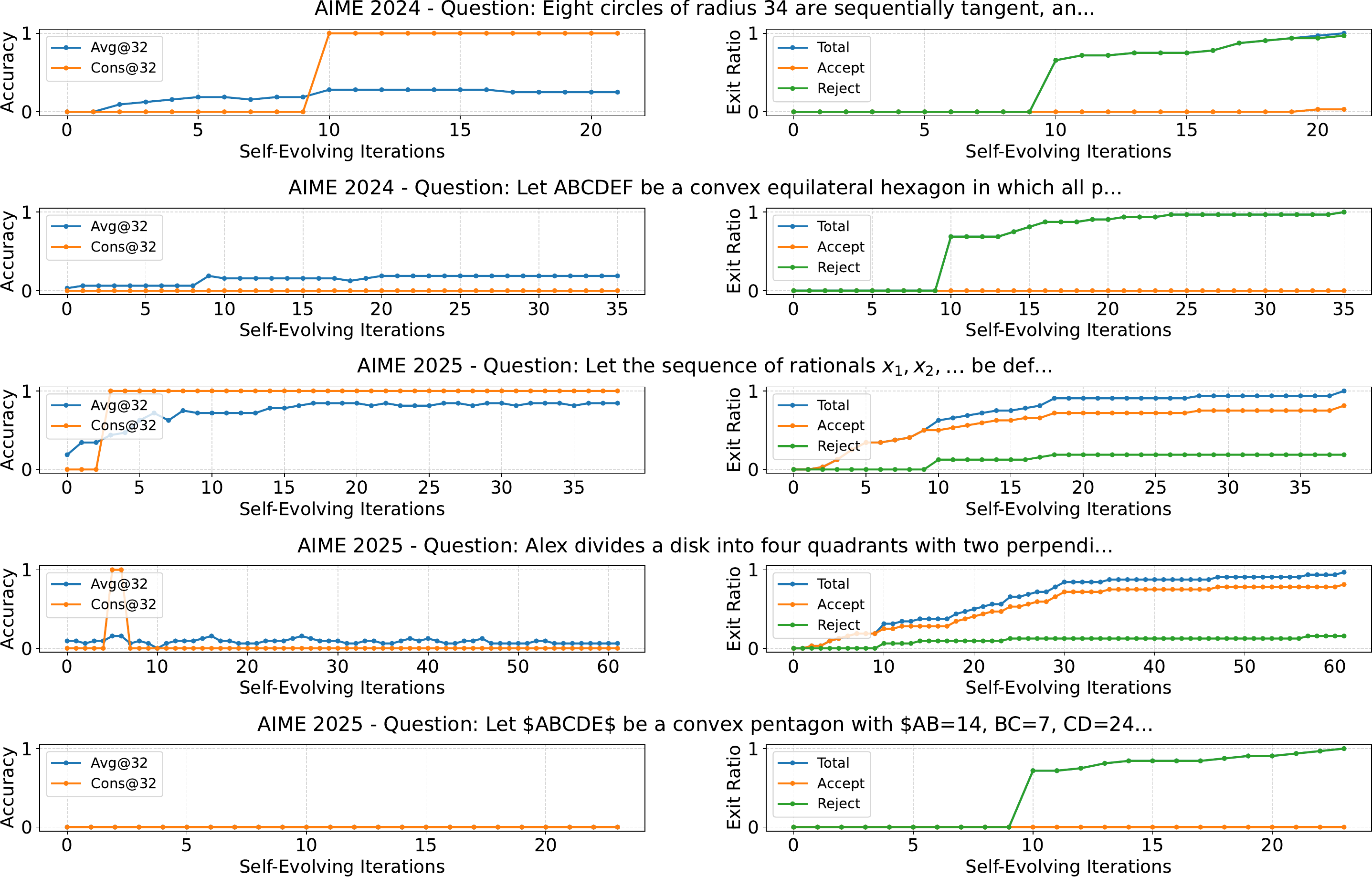}
\caption{
Per-question performance improvements (left) and exit ratios (right) over self-evolving iterations for the ``verification-dependent'' self-evolving approach~\citep{huang2025IMO-Gold-Agent}.
}
\label{fig:imo_agent_acc_on_aime_hard_per_q}
\end{figure*}

\paragraph{Per-Question Convergence Analysis}
Figure~\ref{fig:acc_on_aime_hard_per_q} details the per-question performance improvements for five hard problems ultimately solved by DSER.
We observe very different convergence behaviors and stationary distributions.
For instance, on the top two questions (AIME 2024), DSER leads to quick convergence, and the stationary distribution stabilizes at a high level of solution correctness.
In contrast, for the middle two questions (AIME 2025), convergence is also fast, but the stationary distribution retains a significant portion of incorrect solutions.
For the bottom question (AIME 2025), convergence is very slow, yet DSER eventually achieves the correct majority-voting solution.
These results demonstrate that our approach can successfully leverage different levels of self-improvement capabilities.
Simultaneously, the suboptimal stationary distributions (e.g., the bottom three AIME 2025 questions) highlight the limitations of \texttt{DS-8B} in robustly maintaining correct solutions for certain hard problems.

\paragraph{Comparison with ``Verification-Dependent'' Self-Evolving}
We applied the ``verification-dependent'' self-evolving approach (Section~\ref{sec:method_verify-depend}) to \texttt{DS-8B} on the same 9 hard AIME questions, but it only solved 2 of them.
Figure~\ref{fig:imo_agent_acc_on_aime_hard_per_q} (a side-by-side comparison with Figure~\ref{fig:acc_on_aime_hard_per_q}) shows the corresponding performance on the 5 problems that DSER solved.
The empirical observations align well with our theoretical analysis of this approach's Markov chain.
For problems beyond the model's baseline capacity, its self-verification and self-refinement capabilities are unreliable. This leads to premature rejection exits (rows 2 and 5) or false-positive acceptance exits (row 4).
These results imply that our DSER approach is a more stable and effective method for unlocking the deep reasoning potential of models on tasks beyond their current capacity.
It also points to a distinct possible path for bridging the gap between open-weight reasoning models and leading proprietary models.

\section{Conclusion}
\label{sec:conclu}

We introduced DSER, a probabilistic framework that substantially extends the reasoning boundaries of open-weight models, even when their inherent verification and refinement capabilities are weak. Our core innovation lies in reframing iterative reasoning as a convergent Markov chain, where the long-term guarantee of correctness depends not on flawless step-by-step execution but on a marginal statistical bias towards improvement. This principle allows DSER to unlock the latent potential within smaller models through parallel, long-horizon reasoning trajectories.
Empirically, we demonstrated that DSER enables \texttt{DS-8B} to solve AIME problems that were previously beyond its reach, even rivaling its much larger teacher model. This success demonstrates a promising trade-off between model scale and test-time computation, making powerful reasoning more accessible.

Looking forward, this work opens up several exciting research avenues. First, the limitations in self-verification and refinement exposed by our analysis highlight a critical need for new learning objectives. Future training paradigms could explicitly incentivize robust self-critique and constructive self-correction, moving beyond solely optimizing for final-answer accuracy. Second, the DSER framework itself can be refined; integrating more sophisticated search algorithms or learnable verification modules could enhance its efficiency and success rate. Finally, applying DSER to the exploration phase of reinforcement learning, such as in GRPO, could help discover high-quality reasoning traces for the most challenging problems.

Ultimately, DSER establishes that the path to superior reasoning may lie not only in building larger models but also in designing smarter inference-time processes that guide models to deeply evolve their own thoughts. We believe this paradigm shift towards harnessing test-time computation will be a key driver in the next generation of reasoning systems.


\raggedbottom
\bibliography{main}
\bibliographystyle{iclr2026_conference}

\appendix
\section{Appendix}

\subsection{Theoretical Analysis for Verification-Dependent Self-Evolving}
\label{supp_sec:theory_verify_depend}



\begin{figure*}[h]
\centering
\includegraphics[width=0.6\textwidth]{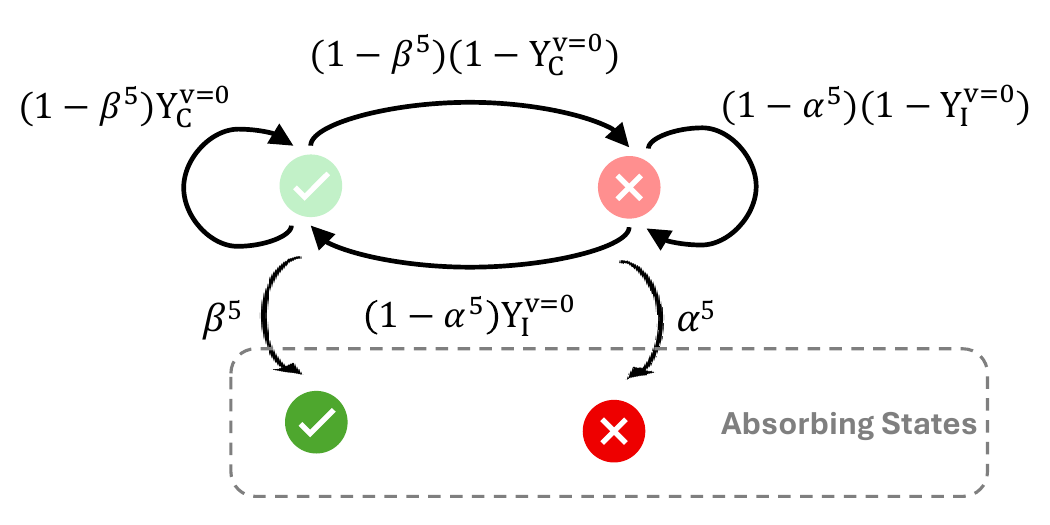}
\caption{
A simplified Markov transition graph for the self-evolving process of~\citep{huang2025IMO-Gold-Agent}, where we remove the rejecting condition of ten consecutive self-verified fails.
}
\label{fig:verify_based_framework}
\end{figure*}

In Section~\ref{sec:method_verify-depend}, we established the self-evolving nature of \citeauthor{huang2025IMO-Gold-Agent}'s framework and analyzed how its absorbing states in the Markov transition graph can prevent deep self-evolution. We now provide a theoretical analysis of a simplified version that removes the rejecting condition, demonstrating that even this variant remains critically dependent on reliable verification capabilities to enable effective self-evolution.
Figure~\ref{fig:verify_based_framework} shows this simplified Markov transition graph.

\paragraph{Markov Transition Model without the Rejecting Condition}
Let $v=0$ and $v=1$ denote a self-verified failure and pass, respectively. Extending the notations defined in Section~\ref{sec:method}, we define the following key conditional probabilities:

$$
\begin{aligned}
\text{In Self-Verification}:\quad&\alpha = p(v=1 \mid X^{(n)}=I), \\
&\beta = p(v=1 \mid X^{(n)}=C), \\
\text{In Self-Refinement}:\quad&Y_I^{v=1} = p(X^{(n+1)}=C \mid X^{(n)}=I, v=1), \\
&Y_C^{v=1} = p(X^{(n+1)}=C \mid X^{(n)}=C, v=1), \\
&Y_I^{v=0} = p(X^{(n+1)}=C \mid X^{(n)}=I, v=0), \\
&Y_C^{v=0}= p(X^{(n+1)}=C \mid X^{(n)}=C, v=0).
\end{aligned}
$$



We define a four-state system to model the process:
\begin{itemize}
    \item State 1 (S1): Correct solution, process ongoing
    \item State 2 (S2): Incorrect solution, process ongoing
    \item State 3 (S3): Correct solution, process terminated (absorbing)
    \item State 4 (S4): Incorrect solution, process terminated (absorbing)
\end{itemize}

S3 and S4 are absorbing-once entered, they transition only to themselves. S3 is reached exclusively from S1 after five consecutive verification passes ($v=1$), while S4 is reached analogously from S2. Transitions between the non-terminated states (S1 and S2) are governed by the refinement probabilities at each iteration. The complete transition probabilities between states are defined as follows:

\[
P = \bordermatrix{
    & S1 & S2 & S3 & S4 \cr
    S1 & (1-\beta^5) Y_C^{v=0} & (1-\beta^5) (1-Y_C^{v=0}) & \beta^5 & 0 \cr
    S2 & (1-\alpha^5) Y_I^{v=0} & (1-\alpha^5) (1-Y_I^{v=0}) & 0 & \alpha^5 \cr
    S3 & 0 & 0 & 1 & 0 \cr
    S4 & 0 & 0 & 0 & 1
}.
\]

\paragraph{Stationary Distribution}

By partitioning the states into absorbing and transient sets, the transition matrix P can be written in the following canonical form:
\[
P = \begin{pmatrix}
Q & R \\
\mathbf{0} & I
\end{pmatrix},\quad
where \quad
Q = \begin{pmatrix}
(1-\beta^5) Y_C^{v=0} & (1-\beta^5) (1-Y_C^{v=0}) \\
(1-\alpha^5) Y_C^{v=0} & (1-\alpha^5) (1-Y_C^{v=0})
\end{pmatrix}, \quad
R = \begin{pmatrix}
\beta^5 & 0 \\
0 & \alpha^5.
\end{pmatrix}
\]
Then we have
$$
P^\infty = \lim_{n\to\infty} P^n = \lim_{n\to\infty} \begin{pmatrix}Q^n & \left(\sum_{k=0}^{n-1} Q^k\right)R \\
\mathbf{0} & I\end{pmatrix} = \begin{pmatrix} \mathbf{0} & (I-Q)^{-1}R \\ \mathbf{0} & I \end{pmatrix}
$$

$$
\begin{aligned}
(I - Q)^{-1}R &= \frac{1}{\det(I - Q)} \begin{pmatrix}
(1 - (1-\alpha^5) (1-Y_I^{v=0}))\beta^5 & (1-\beta^5)\alpha^5 (1-Y_C^{v=0}) \\
(1-\alpha^5)\beta^5 Y_I^{v=0} & (1 - (1-\beta^5) Y_C^{v=0})\alpha^5
\end{pmatrix}, \\
\det(I - Q) &= \alpha^5 - (1-\beta^5) \alpha^5 Y_C^{v=0} + (1-\alpha^5) \beta^5 Y_I^{v=0}
\end{aligned}
$$

The probability of stabilizing in the correct solution is 
$$
\frac{(1-\alpha^5) \beta^5 Y_I^{v=0}}{\alpha^5 - (1-\beta^5) \alpha^5 Y_C^{v=0} + (1-\alpha^5) \beta^5 Y_I^{v=0}}.
$$

\paragraph{Over-Confident Verification Leading to Incorrect Solutions Dominated}
We can prove that under the condition $\alpha^5 \geq Y_I^{v=0}$, which means the problem is difficult and the LLM is over-confident about its solution (a high $\alpha$). In the meanwhile, since the problem is hard, the LLM's capability of making improvements on its solution is limited (a relatively small $Y_I^{v=0}$). The probability of reaching the correct solution will not pass 0.5.



Given $\alpha^5 \geq Y_I^{v=0}$, we have
\begin{align*}
& \frac{1}{\alpha^5} \leq\frac{1}{Y_I^{v=0}} \\
& \frac{1}{\alpha^5} \leq \frac{1 -  Y_C^{v=0}}{ Y_I^{v=0}}+\frac{Y_C^{v=0}}{Y_I^{v=0}} \\
& \frac{1}{\alpha^5} \leq \frac{1}{\beta^5}\frac{1 -  Y_C^{v=0}}{ Y_I^{v=0}}+\frac{Y_C^{v=0}}{Y_I^{v=0}} \\
& \frac{1}{\alpha^5} \leq \frac{1}{\beta^5} \frac{1 -  Y_C^{v=0}}{ Y_I^{v=0}}+\frac{Y_C^{v=0}}{Y_I^{v=0}}+1 \\
\end{align*}

We can calculate that
\begin{align*}
&\frac{(1-\alpha^5) \beta^5 Y_I^{v=0}}{\alpha^5 - (1-\beta^5) \alpha^5 Y_C^{v=0} + (1-\alpha^5) \beta^5 Y_I^{v=0}} \leq \frac{1}{2} \\
\Longleftrightarrow\quad & 2(1-\alpha^5) \beta^5 Y_I^{v=0} \leq \alpha^5 - (1-\beta^5) \alpha^5 Y_C^{v=0} + (1-\alpha^5) \beta^5 Y_I^{v=0} \\
\Longleftrightarrow\quad & (1-\alpha^5) \beta^5 Y_I^{v=0} \leq \alpha^5 - (1-\beta^5) \alpha^5 Y_C^{v=0} \\
\Longleftrightarrow\quad & (1-\alpha^5) \beta^5 Y_I^{v=0} + (1-\beta^5) \alpha^5 Y_C^{v=0} \leq \alpha^5 \\
\Longleftrightarrow\quad & \frac{(1-\alpha^5)}{\alpha^5} \beta^5 Y_I^{v=0} + (1-\beta^5) Y_C^{v=0} \leq 1 \\
\Longleftrightarrow\quad & \frac{(1-\alpha^5)}{\alpha^5} \beta^5 Y_I^{v=0} \leq 1 - (1-\beta^5) Y_C^{v=0} \\
\Longleftrightarrow\quad & \frac{(1-\alpha^5)}{\alpha^5} \leq \frac{1 - (1-\beta^5) Y_C^{v=0}}{\beta^5 Y_I^{v=0}} \\
\Longleftrightarrow\quad & \frac{1}{\alpha^5} -1 \leq \frac{1}{\beta^5} \frac{1 -  Y_C^{v=0}}{ Y_I^{v=0}}+\frac{Y_C^{v=0}}{Y_I^{v=0}} \\
\Longleftrightarrow\quad & \frac{1}{\alpha^5} \leq \frac{1}{\beta^5} \frac{1 -  Y_C^{v=0}}{ Y_I^{v=0}}+\frac{Y_C^{v=0}}{Y_I^{v=0}}+1
\end{align*}

Proof completed.

\subsection{Case Studies in Deep Self-Evolving}
\label{supp_sec:case_study}

To illustrate the inner workings of DSER, we manually trace the critical verification and refinement steps across the 80-iteration process that guided \texttt{DS-8B} to a correct solution for a previously unsolvable AIME 2025 problem.
Below we highlight some crucial verification and refinement actions emerged in DSER.

\begin{casebox}
\sectiontitle{Problem}
 Let $ABCDE$ be a convex pentagon with $AB=14, BC=7, CD=24, DE=13, EA=26,$ and $\angle B=\angle E=60^\circ$. For each point $X$ in the plane, define $f(X)=AX+BX+CX+DX+EX$. The least possible value of $f(X)$ can be expressed as $m+n\sqrt{p}$, where $m$ and $n$ are positive integers and $p$ is not divisible by the square of any prime. Find $m+n+p$

\sectiontitle{Solution}

\texttt{<think>}...\texttt{<\textbackslash think>}

The convex pentagon \(ABCDE\) has side lengths \(AB = 14\), \(BC = 7\), \(CD = 24\), \(DE = 13\), \(EA = 26\), and angles \(\angle B = \angle E = 60^\circ\). The function \(f(X) = AX + BX + CX + DX + EX\) is minimized at a point \(X\) with coordinates \(\left(\frac{109}{7}, \frac{44\sqrt{3}}{7}\right)\).

The distances from \(X\) to the vertices are:
\begin{itemize}
\item \(AX = 5\sqrt{3}\)
\item \(BX = 19\)
\item \(CX = 8\sqrt{3}\)
\item \(DX = 8\sqrt{3}\)
\item \(EX = 19\)
\end{itemize}

The sum is \(f(X) = 5\sqrt{3} + 19 + 8\sqrt{3} + 8\sqrt{3} + 19 = 38 + 21\sqrt{3}\).

This sum is expressed as \(m + n\sqrt{p}\) where \(m = 38\), \(n = 21\), and \(p = 3\). Thus, \(m + n + p = 38 + 21 + 3 = 62\).

The minimum value is achieved, and the sum is consistent with the given conditions.

\boxed{\textcolor{red}{62}}

\sectiondivider

\sectiontitle{Verify}

\texttt{<think>}...\texttt{<\textbackslash think>}

The solution claims that the minimum value of \(f(X) = AX + BX + CX + DX + EX\) occurs at the point \(X = \left(\frac{109}{7}, \frac{44\sqrt{3}}{7}\right)\), with the sum \(f(X) = 38 + 21\sqrt{3}\). However, verification shows that the sum of the unit vectors from the points to \(X\) is not zero, which is a necessary condition for the minimum of the sum of Euclidean distances in the plane for distinct points not collinear. Additionally, calculating \(f(X)\) at another point, such as \((10, 10)\), \textcolor{red}{gives a smaller value (\(f(10, 10) \approx 70.07\)) compared to \(f(X) \approx 74.37\)}, confirming that \(X\) is not the minimum. The pentagon is convex and satisfies the given side lengths and angles, but the minimum sum of distances is not achieved at \(X\). Therefore, the solution is incorrect.

\boxed{0}

\sectiondivider

\sectiontitle{Refinement}

\texttt{<think>}...\texttt{<\textbackslash think>}

The convex pentagon \(ABCDE\) has side lengths \(AB=14\), \(BC=7\), \(CD=24\), \(DE=13\), \(EA=26\), and angles \(\angle B = \angle E = 60^\circ\). The point that minimizes the sum of distances \(f(X) = AX + BX + CX + DX + EX\) is found to be the \textcolor{green}{Fermat-Torricelli point of the triangle formed by vertices \(A\), \(C\), and \(D\)}, which minimizes the sum \(AX + CX + DX\). This sum is \(19\sqrt{3}\), and the minimum value of \(f(X)\) is \(38 + 19\sqrt{3}\), achieved when \(BX + EX = 38\), but this is not possible in this pentagon configuration. However, the lower bound is \(38 + 19\sqrt{3}\), and the minimum value is \(38 + 19\sqrt{3}\).

The expression \(38 + 19\sqrt{3}\) is in the form \(m + n\sqrt{p}\), where \(m = 38\), \(n = 19\), and \(p = 3\). Thus, \(m + n + p = 38 + 19 + 3 = 60\).

\boxed{\textcolor{green}{60}}
\end{casebox}

\begin{casebox}
\sectiontitle{Problem}
 Let $ABCDE$ be a convex pentagon with $AB=14, BC=7, CD=24, DE=13, EA=26,$ and $\angle B=\angle E=60^\circ$. For each point $X$ in the plane, define $f(X)=AX+BX+CX+DX+EX$. The least possible value of $f(X)$ can be expressed as $m+n\sqrt{p}$, where $m$ and $n$ are positive integers and $p$ is not divisible by the square of any prime. Find $m+n+p$
 \sectiondivider

\sectiontitle{Solution}

\texttt{<think>}...\texttt{<\textbackslash think>}

The convex pentagon \(ABCDE\) with side lengths \(AB = 14\), \(BC = 7\), \(CD = 24\), \(DE = 13\), \(EA = 26\), and angles \(\angle B = \angle E = 60^\circ\) is correctly constructed with vertices at \(A(7, 7\sqrt{3})\), \(B(0, 0)\), \(C(7, 0)\), \(D(205/7, 36\sqrt{3}/7)\), and \(E(218/7, 88\sqrt{3}/7)\). The pentagon is convex, as all cross products during traversal are negative, indicating clockwise turns.

The function \(f(X) = AX + BX + CX + DX + EX\) is minimized on the line segment \(BE\), where \(BX + EX = 38\) for all \(X\) on \(BE\), and \(BX + EX > 38\) for \(X\) not on \(BE\). Minimizing \(f(X)\) on \(BE\) is equivalent to minimizing \(AX + CX + DX + 38\). The minimum of \(AX + CX + DX\) on \(BE\) occurs at the midpoint \(M(109/7, 44\sqrt{3}/7)\), where \(AX + CX + DX = 21\sqrt{3}\). Thus, \(f(M) = 38 + 21\sqrt{3}\).

The value \(38 + 21\sqrt{3}\) is the minimum, as \(f(X)\) is larger at other points on \(BE\) and at the vertices, and the minimum occurs on \(BE\) due to the minimization of \(BX + EX\). The expression \(38 + 21\sqrt{3}\) is in the form \(m + n\sqrt{p}\) with \(m = 38\), \(n = 21\), and \(p = 3\) (square-free). Therefore, \(m + n + p = 38 + 21 + 3 = 62\).

\boxed{\textcolor{red}{62}}

\sectiondivider

\sectiontitle{Verify}

\texttt{<think>}...\texttt{<\textbackslash think>}

The solution provided in the query claims that the minimum value of \(f(X) = AX + BX + CX + DX + EX\) is \(38 + 21\sqrt{3}\), achieved at the midpoint M of BE. However, numerical calculations show that the minimum occurs on the line segment BE but at a different point, approximately at parameter t = 0.33, with a value of approximately 70.913. \textcolor{red}{At M, f(X) = 74.372, which is larger than the minimum found.} The minimum value 70.913 cannot be expressed as \(m + n\sqrt{p}\) with integer m, n and square-free p, but the problem requires this form. Since the solution's minimum value and the point are incorrect, the answer is 0.

\boxed{0}

\sectiondivider

\sectiontitle{Refinement}

\texttt{<think>}...\texttt{<\textbackslash think>}

The convex pentagon \(ABCDE\) has vertices \(A(7, 7\sqrt{3})\), \(B(0, 0)\), \(C(7, 0)\), \(D(205/7, 36\sqrt{3}/7)\), and \(E(218/7, 88\sqrt{3}/7)\), with side lengths \(AB = 14\), \(BC = 7\), \(CD = 24\), \(DE = 13\), \(EA = 26\), and angles \(\angle B = \angle E = 60^\circ\). \textcolor{green}{The function \(f(X) = AX + BX + CX + DX + EX\) is minimized on the line segment \(BE\)}, where \(BX + EX = 38\) for all \(X\) on \(BE\), and \(BX + EX > 38\) for \(X\) not on \(BE\). Minimizing \(f(X)\) is equivalent to minimizing \(g(X) = AX + CX + DX + 38\) for \(X\) on \(BE\).

The minimum value of \(f(X)\) is \(38 + 19\sqrt{3}\), achieved at a point on \(BE\). This is expressed as \(m + n\sqrt{p}\) with \(m = 38\), \(n = 19\), and \(p = 3\) (square-free). Thus, \(m + n + p = 38 + 19 + 3 = 60\).

\boxed{\textcolor{green}{60}}
\end{casebox}

\subsection{Additional Experimental Results}
\label{supp_sec:add_exp_res}

Table~\ref{tab:acc_on_hard_aime_q} summarized detailed information of 9 hard problems on AIME 2024-2025 as well as a case-by-case comparison between \texttt{DS-8B} and its enhancement with our DSER approach.

\begin{table}[ht]
\centering
\small
\caption{Performance of \texttt{DS-8B} and its enhancement with our DSER framework on challenging AIME 24/25 problems. 
\textbf{Avg@128} reports the average accuracy of the base model over 128 independent runs. The DSER process is executed independently for 64 runs, and for each run, the results from the final ten self-evolve iterations are aggregated (yielding a total of 640 solutions in total). 
\textbf{Avg@640 (DSER)} thus reflects the average accuracy across all these solutions, while \textbf{Cons@640 (DSER)} indicates whether the majority vote among them yields a correct answer.}
\label{tab:aime_results}
\begin{tabular}{l p{6.5cm} p{1.2cm} p{1.2cm} p{1.2cm}}
\toprule
Source & Question & Avg@128 & Avg@640 (DSER) & Cons@640 (DSER) \\
\midrule
AIME 24 & Let ABCDEF be a convex equilateral hexagon in which all pairs of opposite sides are parallel. The triangle whose sides are extensions of segments AB, CD, and EF has side lengths 200, 240, and 300. Find the side length of the hexagon. (\href{https://artofproblemsolving.com/wiki/index.php/2024_AIME_II_Problems/Problem_5}{url}) & 0.1812 & 0.8016 & 1 \\
AIME 24 & Eight circles of radius $34$ are sequentially tangent, and two of the circles are tangent to $AB$ and $BC$ of triangle $ABC$, respectively. $2024$ circles of radius $1$ can be arranged in the same manner. The inradius of triangle $ABC$ can be expressed as $\frac{m}{n}$, where $m$ and $n$ are relatively prime positive integers. Find $m+n$. (\href{https://artofproblemsolving.com/wiki/index.php/2024_AIME_I_Problems/Problem_8}{url}) & 0.0625 & 0.6531 & 1 \\
AIME 24 & Find the number of rectangles that can be formed inside a fixed regular dodecagon ($12$-gon) where each side of the rectangle lies on either a side or a diagonal of the dodecagon. The diagram below shows three of those rectangles.
(\href{https://artofproblemsolving.com/wiki/index.php/2024_AIME_II_Problems/Problem_15}{url}) & 0.0063 & 0.0016 & 0 \\
AIME 24 & Define $f(x)=|| x|-\tfrac{1}{2}|$ and $g(x)=|| x|-\tfrac{1}{4}|$. Find the number of intersections of the graphs of $y=4 g(f(\sin (2 \pi x))) \quad\text{and}\quad x=4 g(f(\cos (3 \pi y)))$ (\href{https://artofproblemsolving.com/wiki/index.php/2024_AIME_I_Problems/Problem_12}{url}) & 0.0000 & 0.0000 & 0 \\
\midrule
AIME 25 & Let the sequence of rationals $ x_1, x_2, \ldots $ be defined such that $ x_1 = \frac{25}{11} $ and
$ x_{k+1} = \frac{1}{3} \left( x_k + \frac{1}{x_k} - 1 \right). $
$ x_{2025} $ can be expressed as $ \frac{m}{n} $ for relatively prime positive integers $ m $ and $ n $. Find the remainder when $ m + n $ is divided by 1000. (\href{https://artofproblemsolving.com/wiki/index.php/2025_AIME_II_Problems/Problem_13}{url}) & 0.0750 & 0.4016 & 1 \\
AIME 25 & Alex divides a disk into four quadrants with two perpendicular diameters intersecting at the center of the disk. He draws 25 more line segments through the disk, drawing each segment by selecting two points at random on the perimeter of the disk in different quadrants and connecting those two points. Find the expected number of regions into which these 27 line segments divide the disk. (\href{https://artofproblemsolving.com/wiki/index.php/2025_AIME_I_Problems/Problem_13}{url}) & 0.0750 & 0.2000 & 1 \\
AIME 25 & There are exactly three positive real numbers $ k $ such that the function
$ f(x) = \frac{(x - 18)(x - 72)(x - 98)(x - k)}{x} $
defined over the positive real numbers achieves its minimum value at exactly two positive real numbers $ x $. Find the sum of these three values of $ k $. (\href{https://artofproblemsolving.com/wiki/index.php/2025_AIME_II_Problems/Problem_15}{url}) & 0.0375 & 0.0625 & 0 \\
AIME 25 & Let $N$ denote the number of ordered triples of positive integers $(a,b,c)$ such that $a,b,c\leq3^6$ and $a^3+b^3+c^3$ is a multiple of $3^7$. Find the remainder when $N$ is divided by $1000$. (\href{https://artofproblemsolving.com/wiki/index.php/2025_AIME_I_Problems/Problem_15}{url}) & 0.0063 & 0.0000 & 0 \\
AIME 25 & Let $ABCDE$ be a convex pentagon with $AB=14, BC=7, CD=24, DE=13, EA=26,$ and $\angle B=\angle E=60^\circ$. For each point $X$ in the plane, define $f(X)=AX+BX+CX+DX+EX$. The least possible value of $f(X)$ can be expressed as $m+n\sqrt{p}$, where $m$ and $n$ are positive integers and $p$ is not divisible by the square of any prime. Find $m+n+p$. (\href{https://artofproblemsolving.com/wiki/index.php/2025_AIME_I_Problems/Problem_14}{url}) & 0.0000 & 0.2516 & 1 \\
\bottomrule
\end{tabular}
\label{tab:acc_on_hard_aime_q}
\end{table}

\end{document}